%% file: main.tex
\documentclass{article}

\usepackage{arxiv}

\usepackage[utf8]{inputenc} % allow utf-8 input
\usepackage[T1]{fontenc}    % use 8-bit T1 fonts
\usepackage{hyperref}       % hyperlinks
\usepackage{url}            % simple URL typesetting
\usepackage{booktabs}       % professional-quality tables
\usepackage{amsfonts}       % blackboard math symbols
\usepackage{nicefrac}       % compact symbols for 1/2, etc.
\usepackage{microtype}      % microtypography
\usepackage{cleveref}       % smart cross-referencing
\usepackage{lipsum}         % Can be removed after putting your text content
\usepackage{graphicx}
\usepackage{natbib}
\usepackage{doi}

\title{Julearn: an easy-to-use library for leakage-free evaluation and inspection of ML models}

\usepackage{authblk}

\newbox{\orcid}\sbox{\orcid}{\includegraphics[scale=0.06]{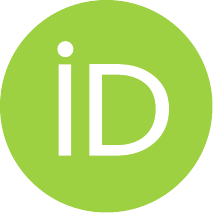}} 
\author[1,2]{
	\href{https://orcid.org/0000-0001-5072-542X}{\usebox{\orcid}\hspace{1mm}Sami Hamdan}}
\author[1,2]{\href{https://orcid.org/0000-0002-1272-217X}{\usebox{\orcid}\hspace{1mm}Shammi More}}
\author[1,2,3]{\href{https://orcid.org/0000-0002-2400-3404}{\usebox{\orcid}\hspace{1mm}Leonard Sasse}}
\author[1,2]{\href{https://orcid.org/0000-0001-9486-0922}{\usebox{\orcid}\hspace{1mm}Vera Komeyer}}
\author[1,2]{\href{https://orcid.org/0000-0002-0289-5480}{\usebox{\orcid}\hspace{1mm}Kaustubh R. Patil}}
\author[1,2]{\href{https://orcid.org/0000-0003-4087-8259}{and \usebox{\orcid}\hspace{1mm}Federico Raimondo\thanks{Corresponding author: \texttt{f.raimondo@fz-juelich.de}}}}
\author[ ]{for the Alzheimer’s Disease Neuroimaging Initiative\thanks{
Data used in preparation of this article were obtained from the Alzheimer’s Disease Neuroimaging Initiative (ADNI) database (adni.loni.usc.edu). As such, the investigators within the ADNI contributed to the design and implementation of ADNI and/or provided data but did not participate in analysis or writing of this report. A complete listing of ADNI investigators can be found at: \url{http://adni.loni.usc.edu/wp-content/uploads/how_to_apply/ADNI_Acknowledgement_List.pdf}
}}

\affil[1]{Institute of Neuroscience and Medicine (INM-7: Brain and Behaviour), Research Centre Jülich, Germany}
\affil[2]{Institute of Systems Neuroscience, Heinrich Heine University Düsseldorf, Germany}
\affil[3]{Max Planck School of Cognition, Stephanstrasse 1a, Leipzig, Germany}

\renewcommand{\shorttitle}{Julearn}

\hypersetup{
pdftitle={Julearn: an easy-to-use library for leakage-free evaluation and inspection of ML models},
pdfsubject={cs.LG, q-bio.NC},
pdfauthor={Sami Hamdan, Shammi More, Leonard Sasse, Vera Komeyer, Kaustubh R. Patil, Federico Raimondo},
pdfkeywords={Machine Learning, Research Software Library, Neuroscience},
}

\include{acronyms.tex}

\begin{document}
\maketitle

\begin{abstract}
\input{abstract.tex}
\end{abstract}

% keywords can be removed

\section{Introduction}
\input{introduction.tex}

\section{Methods}
\input{methods.tex}

\section{Examples}
\input{examples.tex}

\section{Discussion}
\input{discussion.tex}

\input{statements.tex}

\bibliographystyle{unsrtnat}
\bibliography{2023_Julearn.bib}  %%% Uncomment this line and comment out the ``thebibliography'' section below to use the external .bib file (using bibtex) .

\end{document}

%% file: acronyms.tex
\usepackage{acro}

\newcommand{\acro}[3]{
	\DeclareAcronym{#1}{
		short=#2,
		long=#3,
	}
}
	
\acro{ML}{ML}{Machine Learning}

\acro{MRI}{MRI}{Magnetic Resonance Imaging}

\acro{EEG}{EEG}{electroencephalogram}

\acro{CV}{CV}{cross-validation}

\acro{PCA}{PCA}{Principal Component Analysis}

\acro{API}{API}{Application Programming Interface}

\acro{AI}{AI}{Artificial Intelligence}

\acro{CBPM}{CBPM}{Connectome Based Predictive Modelling}

\acro{GMV}{GMV}{Gray Matter Volume}

\acro{IXI}{IXI}{Information eXtraction from Images}

\acro{CAT}{CAT}{Computational Anatomy Toolbox}

\acro{GM}{GM}{Gray Matter}

\acro{WM}{WM}{White Matter}

\acro{CSF}{CSF}{Cerebro-Spinal Fluid}

\acro{GS}{GS}{Global Signal}

\acro{GPR}{GPR}{Gaussian Process Regression }

\acro{RVR}{RVR}{Relevance Vector Regression}

\acro{SVR}{SVR}{Support Vector Regression}

\acro{MAE}{MAE}{Mean Absolute Error}

\acro{ADNI}{ADNI}{Alzheimer’s Disease Neuroimaging Initiative \url{https://adni.loni.usc.edu/}}
\acro{SVM}{SVM}{Support Vector Machine}
\acro{HCP-YA}{HCP-YA}{Human Connectome Project Young-Adult}

\acro{rs-fMRI}{rs-fMRI}{resting-state functional Magnetic Resonance Imaging}

\acro{ICA-FIX}{ICA-FIX}{Independent Component Analysis and FMRIB's ICA-based X-noiseifier}

\acro{FC}{FC}{Functional Connectivity }
\acro{LOO-CV}{LOO-CV}{Leave-One-Out Cross-Validation}

%% file: abstract.tex
\subsection*{Background}
The fast-paced development of machine learning (ML) methods coupled with its increasing adoption in research poses challenges for researchers without extensive training in ML. In neuroscience, for example, ML can help understand brain-behavior relationships, diagnose diseases, and develop biomarkers using various data sources like magnetic resonance imaging and electroencephalography. The primary objective of ML is to build models that can make accurate predictions on unseen data. Researchers aim to prove the existence of such generalizable models by evaluating performance using techniques such as cross-validation (CV), which uses systematic subsampling to estimate the generalization performance. Choosing a CV scheme and evaluating an ML pipeline can be challenging and, if used improperly, can lead to overestimated results and incorrect interpretations. 
\subsection*{Findings}
We created julearn, an open-source Python library, that allow researchers to design and evaluate complex ML pipelines without encountering in common pitfalls.
In this manuscript, we present the rationale behind julearn’s design, its core features, and showcase three examples of previously-published research projects that can be easily implemented using this novel library.
\subsection*{Conclusions}
Julearn aims to simplify the entry into the ML world by providing an easy-to-use environment with built in guards against some of the most common ML pitfalls. With its design, unique features and simple interface, it poses as a useful Python-based library for research projects. 

%% file: introduction.tex
\ac{ML} is fast becoming an indispensable tool in many research fields. It is rapidly gaining increasing importance within neuroscience, where it is used for understanding brain-behavior relationships \cite{wu_challenges_2023}, predicting disease status and biomarker development using diverse data modalities such as \ac{MRI} and \ac{EEG}. Such thriving applications of \ac{ML} are driven by availability of big data and advances in computing technologies. Yet, for domain experts, acquiring relevant \ac{ML} and programming skills remains a significant challenge. This underscores the need for user-friendly software solutions accessible to domain experts without extensive \ac{ML} training. Such solutions would enable them to quickly evaluate \ac{ML} approaches. 

The goal of an \ac{ML} application is to create a model that provides accurate predictions on new unseen data—i.e., a generalizable model. In this context, the goal of a research project is usually to demonstrate that a generalizable model exists for the prediction task at hand. As a single set of samples is usually available, this goal is achieved by assessing the generalization performance by training the model on a subset of the data and testing it on the hold-out test data. If the model performs well on the test data, then the researcher concludes that the prediction task can be solved in a generalizable manner. One of the most prominent approaches to estimate the generalization performance is \ac{CV}. \ac{CV} is a systematic subsampling approach, which trains and tests \ac{ML} pipelines multiple times using independent data splits \cite{varoquaux_assessing_2017}.  The average performance over the splits is taken as an estimate of generalization. To achieve good performance or other aims like data interpretation it is often necessary to perform additional data processing, for example, feature selection. This results in an \ac{ML} pipeline that performs all the needed operations from data manipulations, training and evaluation. Choosing a \ac{CV} scheme and evaluating an \ac{ML} pipeline can be challenging, and if improperly used, it can lead to incorrect results and misguided insights. This underscores the need for user-friendly software solutions accessible to domain experts without in-depth \ac{ML} and programming training.
Problematically, a common outcome of pitfalls is an overestimation of the generalization performance when using \ac{CV}, i.e. models are reported as being more accurate than what they actually are. Here, we highlight two common pitfalls; data leakage and overfitting of hyperparameters.

Data leakage occurs when the separation between the training and test data is not strictly followed. For instance, using all available data in parts of an \ac{ML} pipeline breaks the required separation between training and test data. Such data leakage invalidates the complete \ac{CV} procedure, as information of the testing set is available during training. For example, one might apply a preprocessing step like z-standardization, or \ac{PCA} on the complete dataset before splitting the data. As the preprocessing step is informed about the test data, the later created and transformed training data will reflect the test data as well. Therefore, the learning algorithm can leverage this leaked test set information through the preprocessing and memorize instead of building a predictive model, thus inflating the generalization estimation of \ac{CV}. Most problematically, data leakage can happen in many ways through programming errors or lacking awareness of this danger. 

A similar pitfall can occur when tuning hyperparameters by first observing their test set performance. Hyperparameters are parameters, not learnable by the algorithms themselves, which greatly impact their prediction performance. To tackle this optimization problem, many practitioners repeat a simple \ac{CV} to evaluate test set performance of different hyperparameter combinations. Problematically, both tuning and estimating out-of-sample performance on the same test data breaks the clear distinction of training and testing, as one both optimizes and evaluates the \ac{ML} pipeline on the same test set. Notably, this can happen very quickly over the natural progression of research projects while iterating through ideas of appropriate hyperparameters. The solution to this pitfall is to select the hyperparameters and evaluate the out-of-sample performance in different data splits, which can be achieved by using a nested \ac{CV}. In conclusion, both pitfalls can happen very easily without any malicious intent, through a lack of ML or programming experience. We developed the open-source python package julearn to allow field experts to circumvent these pitfalls by default while training and evaluating ML pipelines. 

While \ac{ML} experts can navigate these and other pitfalls using expert software, such as scikit-learn, domain experts might not always be aware of the pitfalls or how to handle them. This is why we created julearn, to provide an out-of-the box solution, preventing common mistakes, usable by domain experts. Julearn was created to be easy to use, to be accessible for researchers with diverse backgrounds, and to create reproducible results. Furthermore, we engineered julearn so it is easy to extend and maintain, in order to keep up with constantly evolving fields such as neuroscience and medicine. The accessibility and usability aspects of julearn were decided to be at the core, as we aimed to help researchers to apply \ac{ML}. We accomplished this through a careful design of the \ac{API}, comprising only a few simple key functions and classes to both create and evaluate complex \ac{ML} pipelines. Furthermore, we added several utilities that allow investigators to gain a detailed understanding of the resulting pipelines. In order to keep julearn up to date, we built it on top of scikit-learn \cite{pedregosa_scikit-learn:_2012,abraham_machine_2014} and followed common best practices of software engineering like unit testing and continuous integration.

%% file: methods.tex
\subsection{Basic Usage}
Julearn is built on top of scikit-learn \cite{pedregosa_scikit-learn:_2012, abraham_machine_2014}, one of the most influential \ac{ML} libraries in the Python programming language. While scikit-learn provides a powerful interface for programmers to create complex and individualized \ac{ML} pipelines, julearn mainly adds an abstraction layer, providing a simple interface for novice programmers. Note that in contrast to scikit-learn, julearn focusses on so called supervised \ac{ML} tasks, which include any prediction task with known labels while training and evaluating pipelines. Therefore, pipelines in the context of julearn always refer to supervised \ac{ML} pipelines.

To achieve a simple interface for supervised \ac{ML} problems, we implemented a core function called \texttt{run\_cross\_validation} to estimate a models' performance using \ac{CV}. In this function, the user specifies the data, features, target, preprocessing and model name to evaluate as a ML pipeline in a leakage-free cross-validated manner. We chose the popular and simple tabular data structure of pandas' \texttt{DataFrame} \cite{mckinney_data_2010} for both the input data and the output of \texttt{run\_cross\_validation}. This makes preparing the input, inspecting and analyzing the output of julearn simple and transparent.

\begin{figure}[bt!] %% preferably at bottom or top of column
\centering
\includegraphics[width=\linewidth]{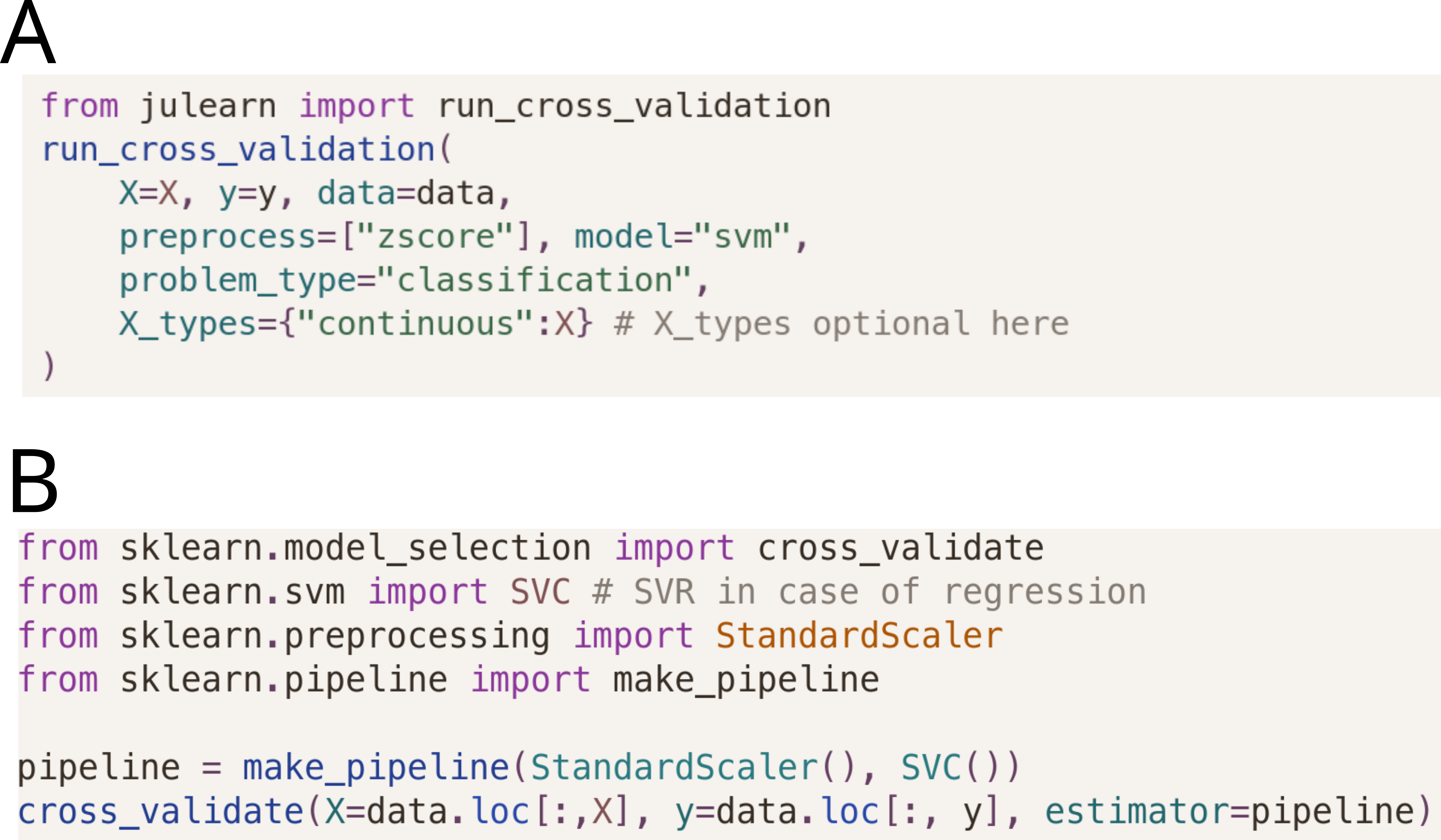}
\caption{Implementation of a simple \ac{CV} pipeline using julearn (A) in contrast to scikit-learn (B). The julearn pipeline needs only one import, while scikit-learn needs multiple. Furthermore, scikit-learn needs to import the Support Vector Machine differently depending on the problem type, while julearn chooses the correct one based on the problem type. The differences between julearn and scikit-learn is most influential for non-experienced programmers who aim to create (complex) supervised \ac{ML} pipelines. Julearn builds upon scikit-learn by providing a simple interface that does not need any awareness of how to compose and find different classes.}\label{fig:cv_julearn_sklearn}
\end{figure}

Furthermore, our \ac{API} provides arguments for feature and target name(s) referring to the columns of the input data frame. To use any of julearn's \ac{ML} algorithms one only needs to provide their name to the \texttt{model} argument of \texttt{run\_cross\_validation}.  Here, julearn will select the model according to the provided \texttt{problem\_type} of either classification or regression. Similarly, one can provide any of the supported preprocessing steps to \texttt{run\_cross\_validation} by name. These steps are executed in a \ac{CV}-consistent way without the risk of data leakage. Such an interface simplifies the construction and use of ML pipelines, in contrast to scikit-learn, where one must import different \ac{ML} models depending on the problem type, create a pipeline using both the imported preprocessing steps and the \ac{ML} model and finally use the \texttt{cross\_validate} function (Figure \ref{fig:cv_julearn_sklearn}).

While julearn does not aim to replace scikit-learn, it tries to simplify specific use cases including the creation of more complex supervised \ac{ML} pipelines that need hyperparameter tuning or preprocessing a subsample of features. This means that julearn can automatically use nested \ac{CV} for proper performance assessment in the context of hyperparameter tuning \cite{poldrack_establishment_2020} and apply preprocessing based on different feature types. These feature types include distinctions like categorical vs continuous features or grouping variables, which can even be used to do confound removal on a subsample of the data.

\subsection{Model Comparison}
In ML applications, there is no standard or consensus of what a good or acceptable performance is, as this usually depends on the task and domain. Thus, the process of developing predictive models involves comparing models, either to null or dummy models, or to previously published models (i.e. benchmarking). Given that \ac{CV} produces estimates of the model's performance and that, depending on the \ac{CV} strategy, these estimates might not be independent from each other, special methods are required to test and conclude if the performance of two models are different or not. For this reason, julearn \texttt{run\_cross\_validation} output has additional information that can be used to do more accurate model comparisons. Furthermore, it provides a stats module, which implements a student's t-test corrected for using the same \ac{CV} approach to compare multiple \ac{ML} pipelines \cite{nadeau_inference_2003}. This correction is necessary as \ac{CV} leads to a dependencency between the folds, i.e., each iteration's training set overlaps with the other ones. To gain a detailed view of the models' benchmark, one can also use julearn's inbuilt visualization tool (see Figure \ref{fig:example_1} for example).

\subsection{Feature Types}
One of the key functionalities that julearn provides that are currently lacking in ML libraries such as scikit-learn is the ability to define feature types. This allows researchers to define sets of variables and do selective processing, needed when dealing with categorical or confounding variables. For this matter, julearn introduces the \texttt{PipelineCreator} to create complex pipelines in which certain processing steps can be applied to one or more subset of features. Once the pipeline is defined, users need to provide a dictionary of any user-defined type and the associated column names in their data as the \texttt{X\_types} argument. Such functionality allows to implement complex pipelines that transform features based on their \textit{type}, e.g., standardizing only continuous features and then deconfound both continuous and categorical features.

\subsection{Hyperparameter Tuning}
\begin{figure}[bt!] %% preferably at bottom or top of column
\centering
\includegraphics[width=\linewidth]{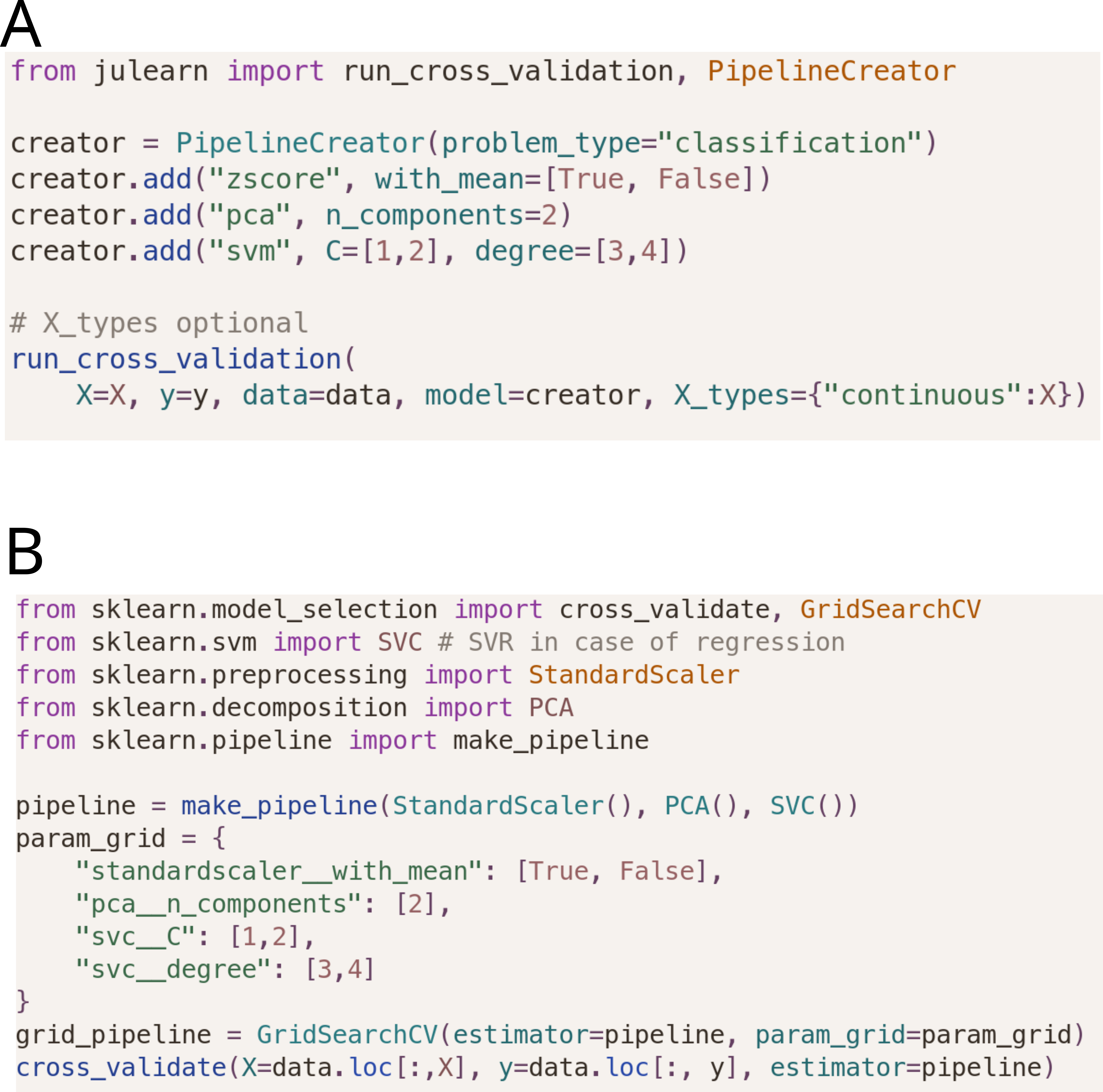}
\caption{Example of julearn (A) and scikit-learn (B) training a typical \ac{ML} pipeline in a \ac{CV} consistent way. Both use a grid search to find optimal hyperparameters. Note that julearn is able to specify the hyperparameters at the same time as it defines each step. On the other hand, scikit-learn needs all hyperparameters to be defined separately with a prefix indicating the step they belong to. This can become complex especially when pipelines are nested, and multiple prefixes are needed.}
\label{fig:pipeline_creator}
\end{figure}

As mentioned previously, hyperparameter tuning should be performed in a nested \ac{CV} to not overfit the predictions of a given pipeline. The \texttt{PipelineCreator} can be used to specify sets of hyperparameters to be tested at each individual step by just using the \texttt{add} method (Figure \ref{fig:pipeline_creator}). Being able to first define a pipeline and its hyperparameters with the \texttt{PipelineCreator}, and to then train and evaluate this pipeline with \texttt{run\_cross\_validation}, makes performing leakage-free nested \ac{CV} easy. In this nested \ac{CV}, all hyperparameters are optimized in an inner \ac{CV} using a grid search by default. This default, like most of julearn's defaults, can be easily adjusted by providing any compatible searcher in the \texttt{run\_cross\_validation}'s \texttt{model\_params} argument. This is a drastic simplification compared to a typical scikit-learn workflow, where one must create the pipeline manually by combining different objects, wrap it inside a \texttt{GridSearchCV} object, and define the hyperparameter options separately from the pipeline itself, using a complex syntax. Lastly, scikit-learn's \texttt{GridSearchCV} object must be provided to its \texttt{cross\_validate} function. 

\subsection{Inspection and Analysis}
Inspection of \ac{ML} pipelines is crucial when working in fields such as neuroscience and medicine as concepts like trustworthy \ac{ML} are heavily dependent on the ability to draw insights and conclusions from models. For this purpose, one needs to be able to inspect and verify each of the pipeline steps, check parameters, evaluate feature importances and further properties of ML pipelines. Julearn includes two functionalities: a \texttt{preprocess\_until} function and \texttt{Inspector} class.  The \texttt{preprocess\_until} function allows users to process the data up to any step of the pipeline, allowing to check how the different transformations are applied. For example, a user might be interested in examining the \ac{PCA} components created or the distribution of features after confound removal. The \texttt{Inspector} object, on the other hand, allows us to inspect the models after estimating their performance using \ac{CV}. It helps users to check fold-wise predictions and obtain both the hyper- and fitted parameters of the trained models. This enables users to verify the robustness of the different parameter combinations and evaluate the variability of the performance across folds. Ongoing efforts to increase julearn's inspection tools encompass integrating tools for explainable \ac{AI} such as SHAP \cite{lundberg_unified_nodate}.

\subsection{Neuroscience-specific Features}
In addition to julearn's field-agnostic features, we also provide neuroscience-specific functionality. Confound removal in the form of confound regression, which is popularly used in neuroscience, was implemented as the \texttt{ConfoundRemover}. This confound regression can be trained on all features or only on specific subsamples defined by a grouping variable, i.e., allowing neuroscientist to only train it on healthy participants as proposed in \citet{dukart_age_2011}. Additionally, we have included the \ac{CBPM} algorithm \cite{shen_using_2017}. This transformer aggregates features significantly correlated with the target into one or two features. This can be done separately for the positively and negatively correlated features. Aggregation can be done using any user specified aggregation function such as summation or mean. We plan to add more neuroscience specific features, such as the integration of harmonization techniques, currently developed in a separate project (\textit{juharmonize}).

\subsection{Customization and Extensibility}
Julearn provides a simple interface to several important \ac{ML} approaches, but it is also easily customizable. Each component of julearn is built to be scikit-learn compatible, meaning that any scikit-learn compatible model and transformer can be provided to \texttt{run\_cross\_validation} and \texttt{PipelineCreator}. Other \texttt{run\_cross\_validation} arguments like \texttt{cv} and hyperparameter searchers were implemented in a way to be extensible by any typical scikit-learn object. This customizability of julearn helps users both extend their usage of julearn and prepares them for the case that they want to transition to scikit-learn to build unique expert level ML pipelines.

%% file: examples.tex
To illustrate the functionality and quality attributes of julearn, we depict three independent examples, showing how the analysis described in previously-published research projects can be implemented with julearn.

\subsection{Example 1: Prediction of age using \ac{GMV} derived from T1-weighted \ac{MRI} images.}

\paragraph{Dataset} We used T1-weighted (T1w) \ac{MRI} images from the publicly available \ac{IXI} dataset (\url{https://brain-development.org/ixi-dataset/}) (IXI, N = 562, age range = 20-86 years) for age estimation similar to \citet{franke_estimating_2010}.

\paragraph{Image Preprocessing} T1w images were preprocessed using the \ac{CAT} version 12.8 \cite{gaser_cat_nodate}. Initial affine registration of T1w images was done with higher than default accuracy (accstr = 0.8), to ensure accurate normalization and segmentation. After bias field correction and tissue class segmentation, accurate optimized Geodesic shooting \cite{ashburner_diffeomorphic_2011} was used for normalization (regstr = 1). We used 1 mm Geodesic Shooting templates and generated 1 mm isotropic images as output. Next, the normalized \ac{GM} segments were modulated for linear and non-linear transformations.

\paragraph{Feature spaces and models}  A whole-brain mask was used to select 238,955 \ac{GM} voxels. Then, smoothing with a 4mm FWHM Gaussian kernel and resampling using linear interpolation to 8 mm spatial resolution was applied resulting in 3747 features.  We tested three regression models \ac{GPR}, \ac{RVR} and \ac{SVR} using this feature space to predict age.

\paragraph{Prediction Analysis} We used 5 times 5-fold \ac{CV} to estimate generalization performance of our pipelines. Hyperparameters were tuned in the inner 5-fold \ac{CV}. Features with low variance were removed (threshold < 1e-5). \ac{PCA} was applied on the features to retain 100\% variance. \ac{GPR} model gave lowest generalization error (mean \ac{MAE} = -5.30 years), followed by \ac{RVR} (\ac{MAE} = -5.56) and \ac{SVR} (\ac{MAE} = -6.98). Corrected t-test revealed significant difference between \ac{GPR} and \ac{SVM} (p = 3.18e-09) and \ac{RVR} and \ac{SVM} (p = 8.19e-09). There was no significant difference between \ac{RVR} and \ac{GPR} (p = 0.075).

\begin{figure}[bt!] %% preferably at bottom or top of column
\centering
\includegraphics[width=\linewidth]{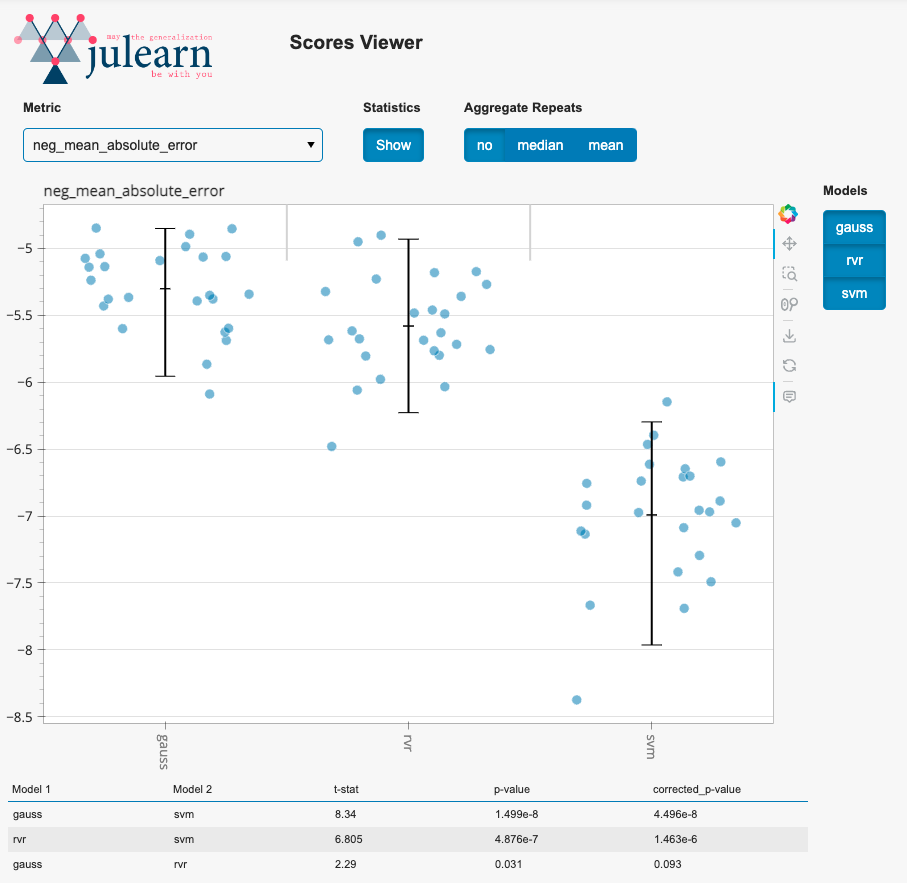}
\caption{Screenshot of the julearn scores viewer, depicting the negative mean absolute error in age prediction from gray matter volume. Each dot represents the negative mean absolute error of each \ac{CV} fold (5 times, 5-folds). Each column represents a different model: \ac{GPR} (gauss), \ac{RVR} (rvr) and \ac{SVR} (svm). Black lines indicate the mean and 95\% confidence intervals. Table at the bottom shows the pairwise statistics using the corrected t-test.}
\label{fig:example_1}
\end{figure}

\subsection{Example 2: Confound Removal.}

\begin{figure}[bt!] %% preferably at bottom or top of column
\centering
\includegraphics[width=\linewidth]{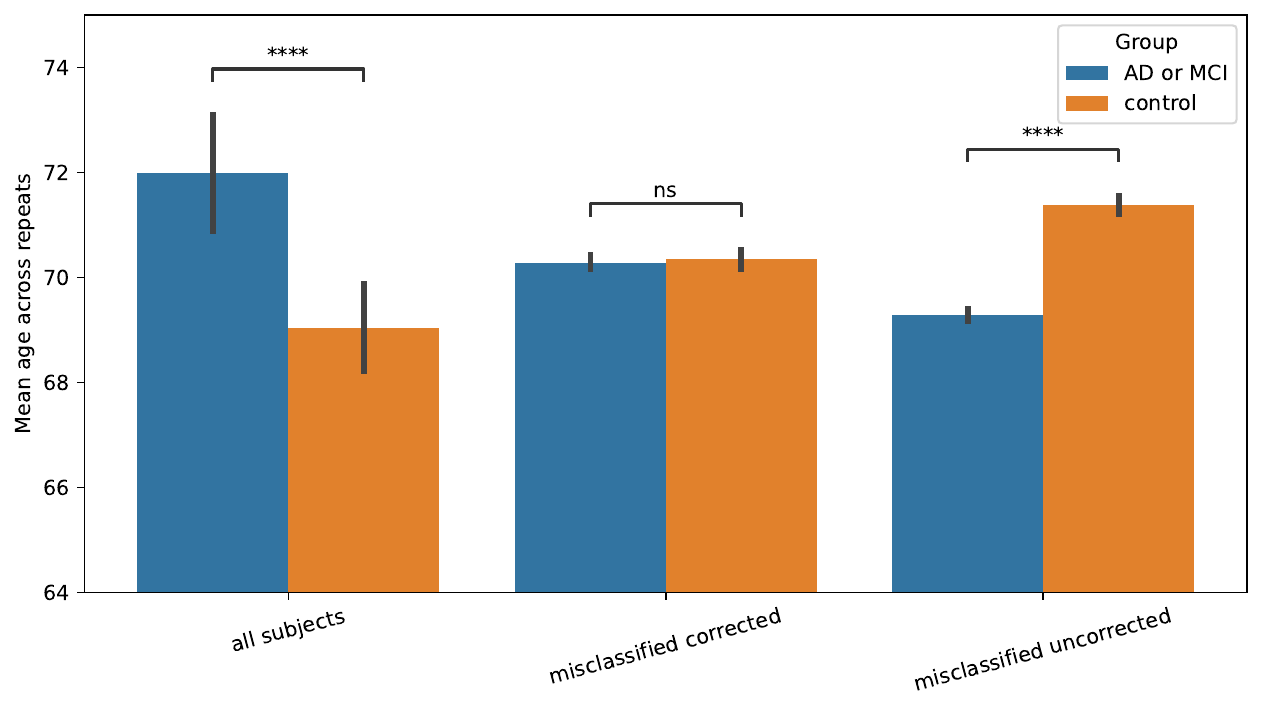}
\caption{Replication of figure 1 in ``Age characteristics of misclassified subjects using \ac{SVM}'' from \citet{dukart_age_2011}. Performing a cross-validated confound removal trained only on the control group using julearn. Julearn greatly simplifies the process to train \ac{CV} consistent preprocessing steps based on characteristics like control vs experimental group. **** means a statistical significance at a p-value threshold of 0.0001 and ns that there is no statistical difference at that threshold.}
\label{fig:example_2}
\end{figure}

\paragraph{Dataset} For this example, we retrieved data conceptually similar to \citet{dukart_age_2011}. We used the \ac{ADNI} database including 498 participants and 68 features. We used age as a confound and the current diagnosis as the target. To simplify the task, we only predict whether a participant has some form of impairment (mild cognitive impairment or Alzheimer's disease) or not (control). 
\paragraph{Prediction Analysis} We aimed to conceptually replicate figure 1 from \citet{dukart_age_2011}. The authors proposed to train confound regression on the healthy participants of a study and then transform all participants using this confound regression. As part of their efforts, they compared two pipelines using the same learning algorithm (\ac{SVM}) \cite{berwick_idiot_1990}. One pipeline was trained to directly classify healthy vs unhealthy participants without controlling for age, while a second pipeline was configured to first control for age using their proposed method: train the confound regression only on healthy participants.  They evaluated the bias of age in the predictions of these models by comparing the age distributions of the healthy vs unhealthy participants for each model’s misclassifications. This was done by computing, for each pipeline, whether there is a significant age difference between these two groups of participants. They found a significant difference when not controlling for age, but not when controlling for age. With further experiments, they conclude that their method leads to less age-related bias. In this example, we replicate the comparison between the two \ac{SVM}s. First, we built both pipelines using julearn and then compared their misclassified predictions to find the same differences (Figure \ref{fig:example_2}).
\\
\\
While the first pipeline (without confound removal) is straightforward to implement, the second variant requires a complicated preprocessing step in which the confound removal needs to be trained on a subsample of one specific column of the data. Thanks to julearn's support for feature types, the whole procedure can be easily implemented by indicating which feature type are to be considered confounds (e.g.: age), which column has the subsampling data (e.g.: current diagnosis) and which values should be considered (e.g.: healthy). Note that the difference between all subjects in age also gets significant for our larger, but not their smaller, sample, which can be attributed to the increased power due to the large sample size. 

\subsection{Example 3: Prediction of fluid intelligence using Connectome-Based Predictive Modelling.}

\paragraph{Dataset} We used data obtained from two \ac{rs-fMRI} sessions from the \ac{HCP-YA} S1200 release \cite{van_essen_wu-minn_2013}. The details regarding collection of behavioral data, \ac{rs-fMRI} acquisition, and image preprocessing have been described elsewhere \cite{glasser_minimal_2013,barch_function_2013}. Here, we provide a an overview. The scanning protocol for \ac{HCP-YA} was approved by the local Institutional Review Board at Washington University in St. Louis. Retrospective analysis of these datasets was further approved by the local Ethics Committee at the Faculty of Medicine at Heinrich-Heine-University in Düsseldorf. We selected sessions for both phase encoding directions (left-to-right [LR] and right-to-left [RL]) obtained on the first day of \ac{HCP-YA} data collection. Due to the \ac{HCP-YA}’s family structure, we selected 399 unrelated subjects (matched for the variable “Gender”), so that we could always maintain independence between folds during cross-validation. In line with \citet{finn_functional_2015}, we filtered out subjects with high estimates of overall head motion (frame-to-frame head motion estimate (averaged across both day 1 rest runs; \ac{HCP-YA}: \textsc{Movement\_RelativeRMS\_mean} > 0.14). This resulted in a dataset consisting of 368 subjects (176 female, 192 male). Participants’ ages ranged from 22 to 37 (M=28.7, SD=3.85). The two sessions of \ac{rs-fMRI} each lasted 15 minutes, resulting in 30 minutes across both sessions. Scans were acquired using a 3T Siemens connectome-Skyra scanner with a gradient-echo EPI sequence (TE=33.1ms, TR=720ms, flip angle = 52°, 2.0mm isotropic voxels, 72 slices, multiband factor of 8).

\paragraph{Image Preprocessing} Data from \ac{rs-fMRI} sessions in the \ac{HCP-YA} had already undergone the HCP’s minimal preprocessing pipeline \cite{glasser_minimal_2013}(, including motion correction and registration to standard space. Additionally, the \ac{ICA-FIX} procedure \cite{salimi-khorshidi_automatic_2014} was applied to remove structured artefacts. Lastly, the 6 rigid-body parameters, their temporal derivatives and the squares of the 12 previous terms were regressed out, resulting in 24 parameters. In addition, we regressed out mean time courses of the \ac{WM}, \ac{CSF}, and \ac{GS}, as well as their squared terms, and the temporal derivatives of the mean signals as well as their squared terms as confounds, resulting in 12 parameters (4 for each noise component). The signal was linearly detrended and bandpass filtered at 0.01 - 0.08 Hz using \texttt{nilearn.image.clean\_img}, The resulting voxel-wise time series were then aggregated using the Shen parcellation \cite{finn_functional_2015} consisting of 268 parcels. \ac{FC} was estimated for each \ac{rs-fMRI} session as Pearson’s correlation between each pair of parcels, resulting in a symmetric 268x268 matrix. These two \ac{FC} matrices were further averaged resulting in one \ac{FC} matrix per subject. One half of the symmetric matrix as well as the diagonal were discarded so that only unique edges were used as features in the prediction workflow.

\begin{figure}[bt!] %% preferably at bottom or top of column
\centering
\includegraphics[width=\linewidth]{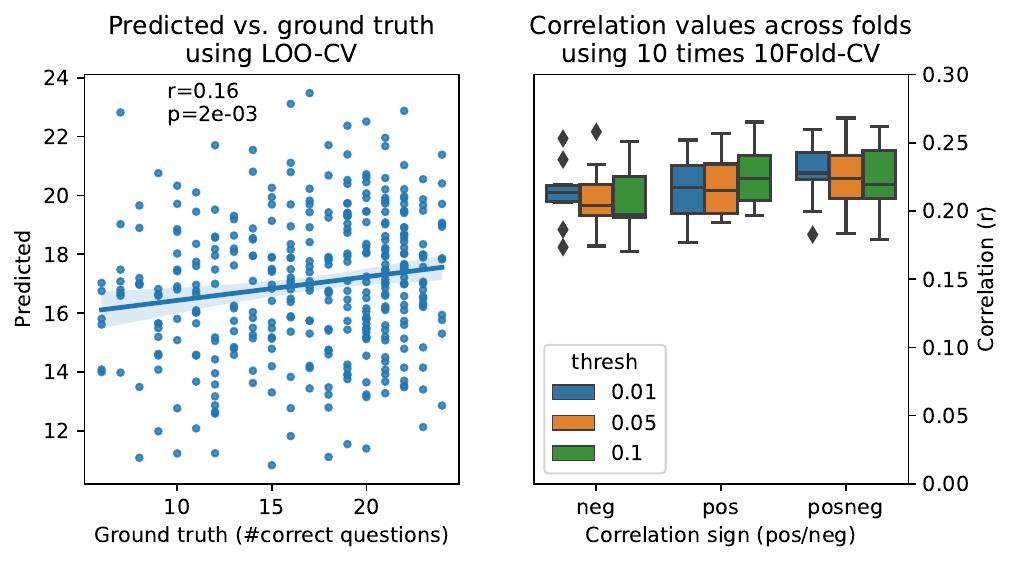}
\caption{results of the prediction of fluid intelligence using \Acl{CBPM} on \ac{HCP-YA} data as in \citet{finn_functional_2015}. Left panel depicts the predicted (y-axis) vs the ground truth (x-axis) values for each sample in a \ac{LOO-CV} scheme, following figure 5a in \citet{finn_functional_2015}. Right panel depicts the mean correlation values (r) across folds, for a 10-times 10-fold \ac{CV} scheme, using different thresholds (colors) and considering either negative correlations, positive correlations, or both kinds of correlations (columns).}
\label{fig:example_3}
\end{figure}

\paragraph{Prediction Analysis} First, we aimed to reproduce \citet{finn_functional_2015} prediction pipeline using the \Acf{CBPM} framework using the \ac{LOO-CV} scheme. Specifically, we reconstructed the workflow used to reproduce Fig 5a in \citet{finn_functional_2015}. As a prediction target we used subjects’ score on the Penn Matrix Test (\textsc{PMAT24\_A\_CR}. This is a non-verbal reasoning assessment and a measure of fluid intelligence. \ac{CBPM} first performs correlation-based univariate feature selection based on a pre-specified significance threshold. Selected features are further divided into positively and negatively correlated features, and then separately summed up, resulting in two features. Subsequently, a linear regression is fitted either on both or one of these features based on user preference. The results here were obtained by using the positive-feature network at a feature selection threshold of p < .01 in line with Fig. 5a from \citet{finn_functional_2015}. We observe a similar trend in our results albeit with a lower correlation between observed and predicted values (see Figure \ref{fig:example_3}). In addition, we also provide results for a 10-Fold cross-validation with 10 repeats. In this analysis, we also tested \ac{CBPM} using positive- and negative-feature networks individually as well as both feature networks combined with varying thresholds for feature selection (0.01, 0.05, 0.1).

%% file: discussion.tex
Julearn aims to bridge the gap between domain expertise in neuroscience and application of \ac{ML} pipelines. Towards that goal, julearn provides a simple interface only using two key \ac{API} points. First, the \texttt{run\_cross\_validation} function provides functionality to evaluate common \ac{ML} pipelines. Second the \texttt{PipelineCreator} provides means to devise complex \ac{ML} pipelines that can be then evaluated using \texttt{run\_cross\_validation}. Additional functionalities are also provided to guide and help users to inspect and evaluate the resulting \ac{CV} scores. In fact, julearn provides a complete workflow for \ac{ML} that has already been used in several publications \cite{mortaheb_mind_2022,more_brain-age_2023}.  Furthermore, the customizability and open-source nature of julearn will help it grow and extend its functionality in the future. 

Julearn does not aim to replace core \ac{ML} libraries such as scikit-learn. Rather its aim is to simplify the entry into the \ac{ML} world by providing an easy-to-use environment with built in guards against some of the most common \ac{ML} pitfalls, such as data leakage that can happen due to not using nested cross-validation and when performing confound removal. Furthermore, julearn is not created to compete with AutoML approaches \cite{ferreira_comparison_2021,zoller_benchmark_2021,waring_automated_2020} which try to automate the preprocessing and modelling over multiple algorithms and sets of hyperparameters. While these approaches are valid and powerful, they yet do not offer full functionality such as nested cross validation and confound removal required in many bio-medical research fields. Furthermore, a researcher might require more control over type of models, parameters and interpretability which might not be easily achievable with current AutoML libraries. Lastly, there are other libraries such as photon\cite{leenings_photonai-python_2021}, Neurominer \cite{noauthor_neurominer_nodate} or Neuropredict \cite{noauthor_neuropredict_nodate} that try to build on top of powerful \ac{ML} libraires to create different interfaces with unique features for field experts. All these libraries are important for a vibrant open-source community, and julearn`s unique features and simple interface will be useful for many research projects.

%% file: statements.tex
\section{Availability of source code}

Julearn’s code is available in GitHub \url{https://github.com/juaml/julearn} with the corresponding documentation in GitHub Pages \url{https://juaml.github.io/julearn/}. The code used for the examples in this manuscript is available at \url{https://github.com/juaml/julearn_paper}, with instructions on how to get the publicly available data.

\begin{itemize}
\item Project name: Julearn
\item Project home page: \url{https://juaml.github.io/julearn/}
\item Operating system(s): Platform independent
\item Programming language: Python
\item License: GNU AGPLv3
\end{itemize}

\section{Data availability}

The data use in this manuscript is publicly available following each dataset requirements. Information on the dataset sources is provided in the description of each example.

\section{Declarations}

\subsection{List of abbreviations}

\printacronyms

\subsection{Consent for publication}

Consent for publication was obtained from the Alzheimer’s Disease Neuroimaging Initiative (ADNI; \url{https://adni.loni.usc.edu/}) Data and Publications Committee. Other datasets do not require consent for publication.

\subsection{Competing Interests}

	The author(s) declare that they have no competing interests'.

\subsection{Funding}

This work was partly supported by the Helmholtz-AI project DeGen (ZT-I-PF-5-078), the Helmholtz Portfolio Theme “Supercomputing and Modeling for the Human Brain” the Deutsche Forschungsgemeinschaft (DFG, German Research Foundation), project PA 3634/1-1 and project-ID 431549029–SFB 1451 project B05, the Helmholtz Imaging Platform and eBRAIN Health (HORIZON-INFRA-2021-TECH-01).

Data collection and sharing for this project was funded by the Alzheimer's Disease Neuroimaging Initiative (ADNI) (National Institutes of Health Grant U01 AG024904) and DOD ADNI (Department of Defense award number W81XWH-12-2-0012). ADNI is funded by the National Institute on Aging, the National Institute of Biomedical Imaging and Bioengineering, and through generous contributions from the following: AbbVie, Alzheimer’s Association; Alzheimer’s Drug Discovery Foundation; Araclon Biotech; BioClinica, Inc.; Biogen; Bristol-Myers Squibb Company; CereSpir, Inc.; Cogstate; Eisai Inc.; Elan Pharmaceuticals, Inc.; Eli Lilly and Company; EuroImmun; F. Hoffmann-La Roche Ltd and its affiliated company Genentech, Inc.; Fujirebio; GE Healthcare; IXICO Ltd.; Janssen Alzheimer Immunotherapy Research \& Development, LLC.; Johnson \& Johnson Pharmaceutical Research \& Development LLC.; Lumosity; Lundbeck; Merck \& Co., Inc.; Meso Scale Diagnostics, LLC.; NeuroRx Research; Neurotrack Technologies; Novartis Pharmaceuticals Corporation; Pfizer Inc.; Piramal Imaging; Servier; Takeda Pharmaceutical Company; and Transition Therapeutics. The Canadian Institutes of Health Research is providing funds to support ADNI clinical sites in Canada. Private sector contributions are facilitated by the Foundation for the National Institutes of Health (www.fnih.org). The grantee organization is the Northern California Institute for Research and Education, and the study is coordinated by the Alzheimer’s Therapeutic Research Institute at the University of Southern California. ADNI data are disseminated by the Laboratory for Neuro Imaging at the University of Southern California.

\subsection{Author's Contributions}

S.H. and F.R. designed the library. S.H, L.S., V.K., S.M., K.R.P. and F.R.. contributed to the development and testing of the library, wrote and reviewed the manuscript. V.K. contributed to the structural design and writing of julearn's documentation. S.M. and F.R. wrote the code for Example 1, S.H. and F.R. wrote the code for Example 2 and L.S. wrote the code for Example 3. 

\section{Acknowledgements}

We want to thank the INM-7 and early adopters of julearn for their valuable contribution at early stages, shaping the direction of our efforts in developing this tool.